# Fuzzy Mixed Integer Optimization Model for Regression Approach

Arindam Chaudhari
Assistant Professor (Computer Science Engineering),
NIIT University, Rajasthan, India
arindam_chau@yahoo.co.in

Dipak Chatterjee
Principal and Professor (Mathematics),
Institute of Engineering and Management, Kolkata,
West Bengal, India

*Abstract*–Mixed Integer Optimization has been a topic of active research in past decades. It has been used to solve Statistical problems of classification and regression involving massive data. However, there is an inherent degree of vagueness present in huge real life data. This impreciseness is handled by Fuzzy Sets. In this Paper, Fuzzy Mixed Integer Optimization Method (FMIOM) is used to find solution to Regression problem. The methodology exploits discrete character of problem. In this way large scale problems are solved within practical limits. The data points are separated into different polyhedral regions and each region has its own distinct regression coefficients. In this attempt, an attention is drawn to Statistics and Data Mining community that Integer Optimization can be significantly used to revisit different Statistical problems. Computational experimentations with generated and real data sets show that FMIOM is comparable to and often outperforms current leading methods. The results illustrate potential for significant impact of Fuzzy Integer Optimization methods on Computational Statistics and Data Mining.

## I. INTRODUCTION

In last few decades, the availability of massive amounts of data in electronic form and significant advances in computational power have given rise to the development of disciplines of Data Mining and Mathematical Programming that is at the centre of modern Scientific and Engineering computation. Two central problems in this direction are data classification and regression. The present focus entails a study of the regression problem. Some of the popular methods for these problems include Decision Trees for classification and regression like CART, C5.0, CRUISE [2], [5], [8], [9], Multivariate Adaptive Regression Splines (MARS) [6] and Support Vector Machines (SVM) [9], [11], [14]. Decision Trees and MARS are heuristic in nature and are closer to statistical inference methods. SVM belongs to the category of separating hyper planes and utilize formal continuous optimization technique like quadratic optimization. These methods are at forefront of Data Mining and have significant impact in practical applications. While Continuous Optimization methods have been widely used in Statistics and have significant impact in last 30 years [1], Integer Optimization has limited impact in Statistical Computing. While statisticians have recognized that problems like classification and regression can be formulated as Integer Optimization problems [1], the belief was formed in early 1970s that these methods are not tractable in practical computational settings. Due to the success of above methods belief of Integer Optimization's impracticality, the applicability of Integer Optimization methods to problems of classification and regression has not been investigated.

Besides this, huge real life data is characterized by an inherent degree of uncertainty and vagueness features. In order to tackle this impreciseness in large data volume Fuzzy Sets serve as an effective tool. Fuzzy theory was originally developed by Zadeh [15] to deal with problems involving linguistic terms [16], [17], [18] and have been successfully applied to various applications in Engineering and Science. It generalizes classical two-valued logic to multi-valued logic for reasoning under uncertainty. Further it is a model-less approach and is clever disguise of Probability Theory. In this Paper, we develop a methodology for regression viz. Fuzzy Mixed Integer Optimization Model (FMIOM) that utilizes Integer Optimization methods using Fuzzy Sets to exploit discrete character of these problems. Due to the significant advances in Integer Optimization in recent past it is possible to solve large scale problems within practical limits. The methodology incorporates clustering to reduce dimensionality, non–linear transformations to improve predictive power, Mixed Integer Optimization to group points together and eliminate outlier data to represent groups by polyhedral regions. The data points are separated into different polyhedral regions and each region has its own distinct regression coefficients. In this attempt, we have drawn the attention of Statistics and Data Mining community that Integer Optimization can be significantly used to revisit different Statistical problems. This Paper is organized as follows. In section II, the geometry of regression approach is illustrated. This is followed by FMIOM for regression in the next section. Computational results and discussions are presented in section IV. Finally, in section V conclusions are given.

## II. GEOMETRY OF REGRESSION APPROACH

In classical regression setting, we have $n$ data points $(x_i, y_i), x_i \in R^d, y_i \in R, i = 1,\ldots,n$. We intend to find a linear relationship between $x_i$ and $y_i$ i.e., $y_i \approx \beta' x_i \forall i$, where the coefficients $\beta \in R^d$ are found by minimizing $\sum_{i=1}^{n}(y_i - \beta' x_i)^2$ or $\sum_{i=1}^{n} | y_i - \beta' x_i |$. In this process $k$ disjoint regions $P_k \subset R^d$ and corresponding coefficients $\beta_k \in R^d, k = 1,........,K$ are found such that if $x_0 \in P_k$, the prediction for $y_0$ will be $\hat{y}_0 = \beta'_r x_0$.

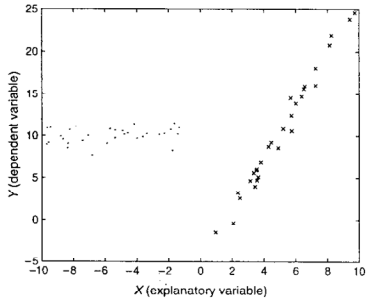

**Figure 1: Set of training data for regression with $d = 1$**

Here, $n$ points are assigned into $K$ groups where, $K$ is user defined parameter. In addition, optimization model is further enhanced to detect and eliminate outlier points in data set (Figure 2). In contrast, traditional regression models deal with outliers after slopes have been determined by examining which points contribute most to total prediction error [12], [13]. This procedure can often be deceiving because the model is heavily influenced by outlier points. After the points are assigned to $K$ groups, we determine coefficients $\beta_k$ that best fit the data for group $k, k = 1,........,K$ and define polyhedra $P_k$ to represent each group using linear optimization methods. After coefficients and polyhedra are defined, we predict $y_0$ value of new point $x_0$. In fact, no partition of $R^d$ are created, so there is a possibility that new point $x_0$ might not belong to any $P_k$. In such a case, we assign it to the region $P_r$ that contains majority among $F$ (a user defined number) nearest neighbors in training set and make prediction $\hat{y}_0 = \beta'_r x_0$. Similarly to the classification model, we preprocess the data by clustering them into small clusters to reduce the dimension and thus computation time of optimization model (Figure 3).

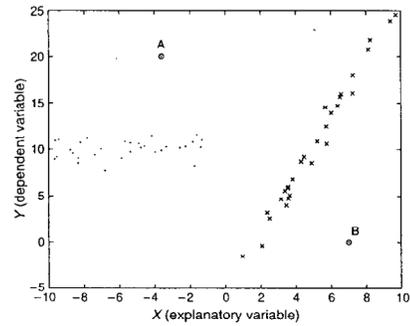

**Figure 2: Outliers in Regression data**

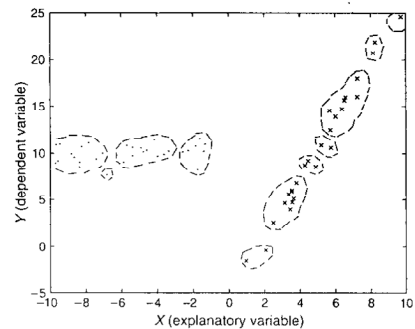

**Figure 3: Clustering of data points**

## III. FMIOM FOR REGRESSION

In this section, we present FMIOM approach for regression. We initiate with Mixed Integer Optimization Method (MIOM) to assign points to groups which is not practical because of dimensionality problems. We first assign points to clusters, then assign clusters to group of points, which are represented by polyhedral regions $P_k$. A method is illustrated to automatically find non-linear transformations of explanatory variables to improve predictive power of method. Finally, the regression algorithm is presented.

*A. Assigning Points to Groups*

The training data consists of $n$ observations $(x_i, y_i), x_i \in R^d, y_i \in R, i = 1,........,n$. We assume $N = \{1,........,n\}, \overline{K} = \{1,........,K\}$ and $M$ as a large positive constant. We define binary variables for $k \in \overline{K}$ and $i \in N$:

$$a_{k,i} = \begin{cases} 1 & x_i \xrightarrow{assigned} k \\ 0 & otherwise \end{cases}$$

The MIOM is as follows:

$$\text{minimize} \sum_{i=1}^{n} \delta_i$$

subject to

$$\delta_i \geq (y_i - \beta_k' x_i) - M(1 - a_{k,i}), k \in \overline{K}, i \in N$$
$$\delta_i \geq -(y_i - \beta_k' x_i) - M(1 - a_{k,i}), k \in \overline{K}, i \in N \quad (1)$$
$$\sum_{k=1}^{K} a_{k,i} = 1, i \in N \ a_{k,i} \in \{0,1\}, \delta_i \geq 0$$

From the first and second constraints $\delta_i$ is absolute error associated with point $x_i$. If $a_{k,i} = 1$, $\delta_i \geq (y_i - \beta_k' x_i)$, $\delta_i \geq -(y_i - \beta_k' x_i)$, and minimization of $\delta_i$ sets $\delta_i$ equal to $|y_i - \beta_k' x_i|$. If $a_{k,i} = 0$, right hand side of first two constraints becomes negative, making them irrelevant because $\delta_i$ is non-negative. Finally, third constraint limits assignment of each point to just one group. It has been found that even for relatively small $n(n \approx 100)$, the above optimization model is difficult to solve in reasonable time. For this reason, a clustering algorithm is executed initially to cluster nearby $x_i$ points together. After $L$ such clusters are found, for $L << n$ we solve FMIOM analogous to above optimization model, but with significantly fewer binary decision variables.

*B. Clustering Algorithm*

Nearest Prototype $(k - NP, k > 1)$ clustering algorithm defined on $R^d$ [4] in $(x, y)$ space is used to find $L$ clusters. The clustering algorithm initiates with $n$ clusters, then continues to merge clusters with points close to each other until $L$ clusters are obtained.

*C. Assigning Points to Groups: Practical Approach*

Continuing with the clustering algorithm of previous section we can find $K$ clusters, define them as our final groups and find the best $\beta_k$ coefficient for each group by solving separate linear regression problems. Such an approach does not combine points to minimize total absolute error. For this reason, we use clustering algorithm until we have $L, L > K$ clusters and then solve MIOM that assigns $L$ clusters into $K$ groups to minimize total absolute error. Another key concern in regression models is the presence of outliers. The MIOM presented next removes potential outliers by eliminating points in clusters that tend to weaken the fit of predictor coefficients.

Let $C_l, l \in \overline{L} = \{1,......,L\}$ be cluster $l$ and denote $l(i)$ as $x_i$'s cluster. Similarly, to optimization problem (1), we define the following binary variables for $k \in \overline{K} \cup \{0\}$ and $l \in \overline{L}$:

$$a_{k,l} = \begin{cases} 1 \ l \xrightarrow{assigned} k \\ 0 \ otherwise \end{cases} \quad (2)$$

We define $k = 0$ as the outlier group in sense that points in cluster $l$ with $a_{0,l} = 1$ will be eliminated. The following fuzzy optimization model assigns clusters to groups and allows possibility of eliminating clusters of points as outliers:

$$\text{minimize} \sum_{i=1}^{n} \tilde{\delta}_i$$

subject to

$$\tilde{\delta}_i \geq (y_i - \tilde{\beta}_k' x_i) - M(1 - a_{k,l(i)}), k \in \overline{K}, i \in N$$
$$\tilde{\delta}_i \geq -(y_i - \tilde{\beta}_k' x_i) - M(1 - a_{k,i}), k \in \overline{K}, i \in N \quad (3)$$
$$\sum_{k=0}^{K} a_{k,l} = 1, l \in \overline{L} \sum_{l=1}^{L} |C_l| a_{0,l} \leq \tilde{\rho} |N|,$$
$$a_{k,l} \in \{0,1\}, \tilde{\delta}_i \geq 0$$

Here, $M$ is a large positive constant and $\tilde{\rho}$ is maximum fraction of points that can be eliminated as outliers. From the first and second set of constraints $\tilde{\delta}_i$ is absolute error associated to point $x_i$. Both $\tilde{\rho}$ and $\tilde{\delta}_i$ are modeled as the following Fuzzy membership function:

$$\mu_{\tilde{x}_j}(x_j) = \begin{cases} \dfrac{d_j}{e_j + d_j - w_j}, w_j < e_j \\ 1, e_j \leq w_j \leq f_j \\ \dfrac{g_j}{w_j - f_j + g_j}, w_j > f_j \end{cases} \quad (4)$$

Here $x$ denotes the variables $\tilde{\rho}$ and $\tilde{\delta}_i$. If $a_{k,l(i)} = 1$, $\tilde{\delta}_i \geq (y_i - \tilde{\beta}_k' x_i)$, $\tilde{\delta}_i \geq -(y_i - \tilde{\beta}_k' x_i)$ and minimization of $\tilde{\delta}_i$ sets it equal to $|y_i - \tilde{\beta}_k' x_i|$. If $a_{k,l(i)} = 0$, the first two constraints become irrelevant because $\tilde{\delta}_i$ is non-negative. The third set of constraint limits assignment of each cluster point to just one group including the outlier group. The last constraint limits percentage of points eliminated to be less than or equal to pre-specified number $\tilde{\rho}$. If $a_{k,l}=1$ then all points in cluster $l$ are assigned to group $k$ i.e. $G_k = \cup_{\{l|a_{k,l}=1\}} C_l$. Optimization Problem (3) has $KL$ variables as opposed to $Kn$ variables in Problem (2). The number of clusters $L$ controls trade-off between the quality of solution and efficiency of computation. As $L$ increases, the quality of solution increases but the efficiency of computation decreases.

*D. Assigning Groups to Polyhedral Regions*

We identify $K$ groups of points solving Optimization Problem (3). In this section, we establish a geometric representation of group $k$ by a polyhedron $P_k$. It is possible for convex hulls of $K$ groups to overlap and thus we might not be able to define disjoint regions of $P_k$ that contain all points of group $k$. For this reason our approach is based on separating pairs of groups with the objective of minimizing sum of violations. We first outline how to separate group $k$ to group $r$, $k < r$. We consider following two Fuzzy Linear Optimization Problems:

$$\min imize \sum_{i \in G_k} \tilde{\varepsilon}_i + \sum_{l \in G_r} \tilde{\varepsilon}_l$$

*subject to*

$$p'_{k,r} x_i - q_{k,r} \leq -1 + \tilde{\varepsilon}_i, i \in G_k \quad (5)$$
$$p'_{k,r} x_l - q_{k,r} \geq 1 - \tilde{\varepsilon}_l, l \in G_r$$
$$p'_{k,r} e \geq 1, \tilde{\varepsilon}_i \geq 0, \tilde{\varepsilon}_l \geq 0$$

$$\min imize \sum_{i \in G_k} \tilde{\varepsilon}_i + \sum_{l \in G_r} \tilde{\varepsilon}_l$$

*subject to*

$$p'_{k,r} x_i - q_{k,r} \leq -1 + \tilde{\varepsilon}_i, i \in G_k \quad (6)$$
$$p'_{k,r} x_l - q_{k,r} \geq 1 - \tilde{\varepsilon}_l, l \in G_r$$
$$p'_{k,r} e \leq -1, \tilde{\varepsilon}_i \geq 0, \tilde{\varepsilon}_l \geq 0$$

Here $e$ is vector of ones, $\tilde{\varepsilon}_i$ and $\tilde{\varepsilon}_l$ are modeled as Fuzzy membership function given by equation (4). Both problems (5) and (6) find a hyper plane $p'_{k,r} x = q_{k,r}$ that softly separates points in group $k$ from points in group $k$. The third constraint prevents the trivial hyper plane $p_{k,r} = 0$ and $q_{k,r} = 0$ for optimal solution. Problems (5) and (6) set the sum of elements of $p_{k,r}$ to be strictly positive and negative respectively. On solving problems (5) and (6) for every pair of groups, we assume

$$P_k = \{ x \mid p'_{k,i} x \leq q_{k,i}, i = 1, \ldots, k-1,$$
$$p'_{k,i} x \geq q_{k,i}, i = k+1, \ldots, K \} \quad (7)$$

After $P_k$ is defined, we re-compute $\beta_k$ using all points contained in $P_k$ because it is possible that they are different from original $G_k$ that problem (7) obtained. The Optimization Problem is solved that minimizes absolute deviation of all points in $P_k$ to find new $\beta_k$.

*E. Non linear Data Transformations*

To improve the predictive power of FMIOM, we augment explanatory variables with non-linear transformations. In particular, we consider transformations $x^2, \log x$ and $1/x$ applied to the co-ordinates of given points. We augment each $d$ dimensional vector $x_i = (x_{i,1}, \ldots, x_{i,d})'$ with $x_{i,j}^2, \log x_{i,j}, 1/x_{i,j}, j = 1, \ldots, d$ and apply FMIOM to resulting $4d$ dimensional vectors, but the increased dimension slows computation time. For this reason simple heuristic method is used to choose which transformation of which variable to include in data set.

*F. Regression Algorithm*

The regression algorithm comprises of following steps: (a) Nonlinear Transformation: Augment original data set with non-linear transformations using method discussed in section III (E); (b) Preprocessing: Use Fuzzy clustering algorithm to find $L \ll n$ clusters of data points; (c) Assign clusters to groups: Solve Optimization Problem (3) to determine which points belong to which group while eliminating potential outliers; (d) Assign groups to polyhedral regions: Solve linear Optimization Problems (5) and (6) for all pairs of groups and define polyhedra as in Equation (7). (e) Re-computation of $\beta$: Once polyhedra $P_k$ are identified, re-compute $\beta_k$ using only the points that belong in $P_k$. Given a new point $x_0$ (augmented by same transformations as applied in training set data), $x_0 \in P_k$ then we predict $\hat{y}_0 = \tilde{\beta}'_r x_0$. Otherwise, we assign $x_0$ to region $P_r$ that contains majority among its $F$ neighboring points in training set and make prediction $\hat{y}_0 = \tilde{\beta}'_r x_0$.

## IV. COMPUTATIONAL RESULTS AND DISCUSSIONS

In this section, we discuss the performance of FMIOM on three real data sets viz. Boston, Abalone and Auto data and Friedman's [7] generated data sets. A comparison is also made on performances of Linear Least Square Regression (LLSR) and Artificial Neural Networks (ANN) with Radial Basis Function (RBF) and generalized regression using MATLAB's ANN Toolbox. Each regression data was split into three parts with 50%, 30% and 20% of data used for training, validation and testing respectively. The assignment to each set was done randomly and the process was repeated 10 times. The validation set was used to fine tune value of parameter $K$. In all cases, the FMIOM depicted in Equation (17) was solved and parameters $M$, $L$ and $\tilde{\rho}$ were set to 10000, 10 and 0.01 respectively. In ANN, validation set was used to select the appropriate model viz. RBF against generalized regression, adjust number of epochs, number of layers, spread constant and accuracy parameter.

Tables I and II illustrate mean absolute error and mean squared error respectively of LLSR, ANN and FMIOM averaged over 10 random partitions on Friedman's data sets. Tables III and IV illustrate mean absolute error and mean squared error respectively of regression methods averaged over 10 random partitions on Boston, Abalone and Auto data sets. The numbers in parenthesis are corresponding standard deviations. Tables V and VI illustrate average running time in CPU seconds of the methods for Friedman's generated data sets and real data sets respectively. We measure the performance of different regression methods by their predictive ability and stability of their solutions. The prediction accuracy are measured using both mean absolute errors and mean squared errors between predicted against the actual response variable in testing set, given that mean absolute error is used as goodness of fit criterion for FMIOM and mean squared error is used as goodness of fit criterion for LLSR and ANN model. FMIOM used $K = 2$ for all data sets, ANN always used RBF as preferred model with just one layer of nodes. Table 6 shows that FMIOM has relatively reasonable average running time as other methods for small data sets, but its run time explodes for larger Abalone data set. There exists a dramatic increase in run time with larger data sets mainly due to $M$ parameter in models (1) and (3). Because tight estimate of $big - M$ parameter cannot be determined apriori, the large value of $M$ seriously hampers the efficiency of FMIOM.

**TABLE I: Mean absolute error of LLSR, ANN and FMIOM on Friedman data sets**

| Real Data | | LLSR | | | ANN | | | FMIOM | | |
|---|---|---|---|---|---|---|---|---|---|---|
| Friedman Data | $n$ | Train | Validation | Test | Train | Validation | Test | Train | Validation | Test |
| F1 | 500 | 0.890 | 0.954 | 0.913 | 0.875 | 0.918 | 0.887 | 0.869 | 0.937 | 0.912 |
| F2 | 500 | 0.939 | 0.963 | 1.034 | 0.923 | 0.942 | 1.011 | 0.914 | 0.967 | 1.031 |
| F3 | 500 | 0.905 | 0.960 | 0.949 | 0.895 | 0.940 | 0.929 | 0.887 | 0.969 | 0.960 |
| F4 | 1000 | 0.944 | 0.941 | 0.966 | 0.938 | 0.931 | 0.957 | 0.931 | 0.942 | 0.964 |
| F5 | 1000 | 0.903 | 0.914 | 0.931 | 0.805 | 0.898 | 0.913 | 0.886 | 0.914 | 0.927 |
| F6 | 1000 | 0.931 | 0.939 | 0.940 | 0.936 | 0.936 | 0.935 | 0.921 | 0.944 | 0.944 |
| F7 | 4000 | 0.951 | 0.965 | 0.964 | 0.948 | 0.959 | 0.960 | 0.946 | 0.960 | 0.961 |
| F8 | 4000 | 0.944 | 0.964 | 0.958 | 0.940 | 0.959 | 0.953 | 0.937 | 0.960 | 0.956 |
| F9 | 4000 | 0.953 | 0.950 | 0.947 | 0.951 | 0.949 | 0.944 | 0.950 | 0.950 | 0.946 |

**TABLE II: Mean squared error of LLSR, ANN and FMIOM on Friedman data sets**

| Real Data | | LLSR | | | ANN | | | FMIOM | | |
|---|---|---|---|---|---|---|---|---|---|---|
| Friedman Data | $n$ | Train | Validation | Test | Train | Validation | Test | Train | Validation | Test |
| F1 | 500 | 1.299 | 1.485 | 1.370 | 1.245 | 1.389 | 1.303 | 1.312 | 1.417 | 1.354 |
| F2 | 500 | 1.353 | 1.428 | 1.619 | 1.300 | 1.355 | 1.544 | 1.346 | 1.444 | 1.617 |
| F3 | 500 | 1.286 | 1.466 | 1.498 | 1.245 | 1.401 | 1.442 | 1.300 | 1.496 | 1.531 |
| F4 | 1000 | 1.446 | 1.426 | 1.515 | 1.426 | 1.403 | 1.489 | 1.450 | 1.444 | 1.512 |
| F5 | 1000 | 1.342 | 1.378 | 1.416 | 1.295 | 1.334 | 1.356 | 1.317 | 1.375 | 1.370 |
| F6 | 1000 | 1.389 | 1.421 | 1.423 | 1.387 | 1.401 | 1.396 | 1.400 | 1.437 | 1.437 |
| F7 | 4000 | 1.442 | 1.471 | 1.482 | 1.431 | 1.455 | 1.469 | 1.437 | 1.460 | 1.470 |
| F8 | 4000 | 1.427 | 1.498 | 1.482 | 1.412 | 1.485 | 1.469 | 1.417 | 1.490 | 1.472 |
| F9 | 4000 | 1.439 | 1.432 | 1.427 | 1.433 | 1.428 | 1.417 | 1.440 | 1.427 | 1.427 |

TABLE III: Mean absolute error of LLSR, ANN and FMIOM on Boston, Abalone and Auto data sets

| Real Data | | | LLSR | | | ANN | | | FMIOM | | |
|---|---|---|---|---|---|---|---|---|---|---|---|
| Data Set | $n$ | $d$ | Train | Validation | Test | Train | Validation | Test | Train | Validation | Test |
| Boston | 506 | 13 | 3.278 | 3.424 | 3.489 | 2.023 | 2.808 | 2.884 | 2.119 | 2.604 | 2.619 |
| Abalone | 4177 | 7 | 1.619 | 1.642 | 1.667 | 1.572 | 1.598 | 1.611 | 1.475 | 1.509 | 1.531 |
| Auto | 392 | 7 | 2.532 | 2.533 | 2.601 | 1.673 | 2.494 | 2.698 | 1.850 | 2.050 | 2.086 |

TABLE IV: Mean squared error of LLSR, ANN and FMIOM on Boston, Abalone and Auto data sets

| Real Data | | | LLSR | | | ANN | | | FMIOM | | |
|---|---|---|---|---|---|---|---|---|---|---|---|
| Data Set | $n$ | $d$ | Train | Validation | Test | Train | Validation | Test | Train | Validation | Test |
| Boston | 506 | 13 | 23.374 | 27.778 | 26.032 | 7.812 | 16.534 | 16.442 | 10.666 | 13.500 | 14.119 |
| Abalone | 4177 | 8 | 4.959 | 5.316 | 5.291 | 4.778 | 4.941 | 4.942 | 4.569 | 4.664 | 4.786 |
| Auto | 392 | 7 | 11.437 | 11.334 | 12.057 | 5.052 | 12.257 | 15.988 | 7.210 | 8.637 | 9.600 |

TABLE V: Average CPU time of LLSR, ANN and FMIOM on Friedman data sets

| Data | LLSR | ANN | FMIOM |
|---|---|---|---|
| F1 | 0 | 0.519 | 4.386 |
| F2 | 0 | 0.146 | 4.310 |
| F3 | 0 | 0.170 | 4.620 |
| F4 | 0 | 0.298 | 37.37 |
| F5 | 0.001 | 0.290 | 46.31 |
| F6 | 0 | 0.284 | 54.37 |
| F7 | 0.001 | 4.711 | 2.370 |
| F8 | 0.001 | 4.702 | 2.031 |
| F9 | 0.001 | 4.672 | 2.182 |

TABLE VI: Average CPU time of LLSR, ANN and FMIOM on Boston, Abalone and Auto data sets

| Data Set | LLSR | ANN | FMIOM |
|---|---|---|---|
| Boston | 0.000 | 1.092 | 0.537 |
| Abalone | 0.000 | 13.103 | 137.37 |
| Auto | 0.000 | 0.824 | 0.131 |

The computational experiments illustrated some benefits and shortcomings of FMIOM compared to other existing methods in Data Mining and Machine Learning. The results obtained from FMIOM can be improved further by enforcing continuity in boundaries, finding stronger approximations of parameter $M$ and making more computational runs. Its main weakness arises from discontinuity of regression line at boundaries of polyhedral regions. FMIOM predictive performance of general continuous function is significantly hampered as evident with Friedman data set. Continuity can be imposed by modifications to the FMIOM in one dimensional case, but extension to higher dimensions is not evident with current model. Thus, as it currently stands if underlying function is continuous, perhaps a continuous model would be more appropriate. Another apparent challenge for FMIOM is maintaining reasonable computation time. Compared to heuristic based methods in Data Mining such as ANN and Classification Trees, FMIOM has much longer running time for larger data sets. However, FMIOM did have faster running times compared to other techniques like SVM in certain problems. The implementation of FMIOM can be improved to speed up its running time. It can implement a tailored quadratic programming solver for solving SVM sub-problems as done in all implementations of SVM. The FMIOM for regression can be tailored by implicitly branching on integer variables [3]. Such an implementation will also eliminate the need for $big-M$ constraints in Optimization Problems (1) and (3) that can significantly hamper computation time of Integer Programming problems. In addition, because provably optimal solution is not critical in this context, we can prematurely terminate branch and bound procedure at 5% and 10% relative optimality gap or by time limit. However, even with all these improvements, FMIOM does not have superior performance over methods like Classification Trees with respect to time. This method would be appropriate for those who value prediction accuracy over computation time, which might be in areas of Medical and Genetic research. Third shortcoming of FMIOM is its lack of interpretability such as ANOVA interpretation. This weakness is shared by ANN and SVM. Unfortunately, not much can be done to improve this problem for FMIOM. Thus, if decision rules or variable importance information are vital to Data Mining application, tools such as Classification Trees would be more suitable. FMIOM may find similar audience that might find its classification accuracy more valuable. Also, FMIOM is able to handle categorical variables like SVM which might be an additional benefit in certain applications.

## V. CONCLUSION

FMIOM presents a new approach to solve Regression problem. The methodology exploits discrete character of problem and incorporates clustering to reduce dimensionality, non–linear transformations to improve predictive power, Mixed Integer Optimization to group points together and eliminate outlier data to represent groups by polyhedral regions. In this way large scale problems are solved within practical limits. The data points are separated into different polyhedral regions. Each region has its own distinct regression coefficients. Computational results on real data sets are encouraging because FMIOM has outperformed other techniques like LLSR and ANN. We hope that these encouraging results will motivate Statistics, Data Mining and Machine Learning community to reexamine Integer Optimization as viable tool in Statistical Computing.